\documentclass[letterpaper, 10 pt, conference]{ieeeconf}  

\IEEEoverridecommandlockouts                              

\usepackage{graphics}
\usepackage{epsfig}
\usepackage{color}
\usepackage{verbatim}
\usepackage{graphicx}
\usepackage{indentfirst}
\usepackage{amsmath}
\usepackage{subfigure}
\usepackage{multirow}
\usepackage{longtable}
\usepackage{float}
\usepackage{url}

\usepackage{graphicx}
\usepackage{subfigure}

\title{\LARGE \bf
Monocular Outdoor Semantic Mapping with a Multi-task Network
}

\author{Yucai Bai$^{1}$, Lei Fan$^{2}$, Ziyu Pan$^{2}$ and Long Chen$^{2*}$
	\thanks{The corresponding author is Long Chen (chenl46@mail.sysu.edu.cn)}
	\thanks{$^{1}$Yucai Bai is with College of Computer Science, Sichuan University, Chengdu, Sichuang, P.R.China. This work was performed when Yucai Bai was visiting Sun Yat-sen University as a visiting student.}%
	\thanks{$^{2}$Lei Fan, Ziyu Pan and Long Chen are with School of Data and Computer Science, Sun Yat-sen University, Guangzhou, Guangdong, P.R.China.}%
}

\begin{document}

\maketitle
\thispagestyle{empty}
\pagestyle{empty}

\begin{abstract}

In many robotic applications, especially for the autonomous driving, understanding the semantic information and the geometric structure of surroundings are both essential. Semantic 3D maps, as a carrier of the environmental knowledge, are then intensively studied for their abilities and applications. However, it is still challenging to produce a dense outdoor semantic map from a monocular image stream. Motivated by this target, in this paper, we propose a method for large-scale 3D reconstruction from consecutive monocular images. First, with the correlation of underlying information between depth and semantic prediction, a novel multi-task Convolutional Neural Network (CNN) is designed for joint prediction. Given a single image, the network learns low-level information with a shared encoder and separately predicts with decoders containing additional Atrous Spatial Pyramid Pooling (ASPP) layers and the residual connection which merits disparities and semantic mutually. To overcome the inconsistency of monocular depth prediction for reconstruction, post-processing steps with the superpixelization and the effective 3D representation approach are obtained to give the final semantic map. Experiments are compared with other methods on both semantic labeling and depth prediction. We also qualitatively demonstrate the map reconstructed from large-scale, difficult monocular image sequences to prove the effectiveness and superiority.

\end{abstract}

\section{Introduction}

Urban environment perception, as one of the core issues for autonomous driving and other road scene related applications, provides valuable information for further localization inferring, obstacle avoidance~\cite{chen2017moving}, drivable area extraction~\cite{li2014sensor} and etc. To successfully understand the 3D world, various methods have been developed for scene parsing and 3D structure acquiring based on visual sensors. The dense semantic map is a suitable presentation method to provide necessary information. However, compared to stereo vision-based approaches, producing a large-scale dense semantic map remains challenging and computation-consuming. In this paper, we propose a method with a new multi-task semantic and depth prediction model and a superpixel-based refinement to overcome these limitations for monocular semantic mapping.

The current leading scene parsing methods, such as the PSPNet~\cite{zhao2017pyramid} and the Deeplab~\cite{chen2018encoder}, are designed to assign a semantic label to each pixel. The semantic segmentation methods are mostly based on the encoder-decoder network. The architecture first encodes the input images into concentrated features and then decodes to potentials belonging to each semantic category. 
For the estimation of depth from only one image, recent works~\cite{eigen2014depth,laina2016deeper,kumar2018monocular,zhou2017unsupervised} deal with this issue as a parametric learning process. The leading approaches using the CNNs can predict disparities from a single image based on strong fitting abilities and gain significant accuracy.

In particular, the two tasks, i.e., semantic labeling and depth prediction, are strongly interrelated to each other and separate computations bring redundancies. A popular paradigm for predicting different labels is to leverage a multi-task network~\cite{kendall2017multi,neven2017fast} which combines multiple loss functions to learn diverse objectives. Some specific features that are hard to capture for one task but easy for another can be efficiently obtained within multi-task networks. The risk of overfitting is also reduced with the multi-task training process~\cite{caruana1997multitask}. In order to explore the potential relationship between depth and semantic knowledge while reducing the computation cost, we design a network with a shared encoder and two decoders to predict depth and semantic information from a single image simultaneously. 
The output depth is benefited from the connection between two decoders. We also derive the ASPP module in both decoders to extend the fields-of-view.

\begin{figure}[t]
	\centering
	\includegraphics[width=0.48\textwidth]{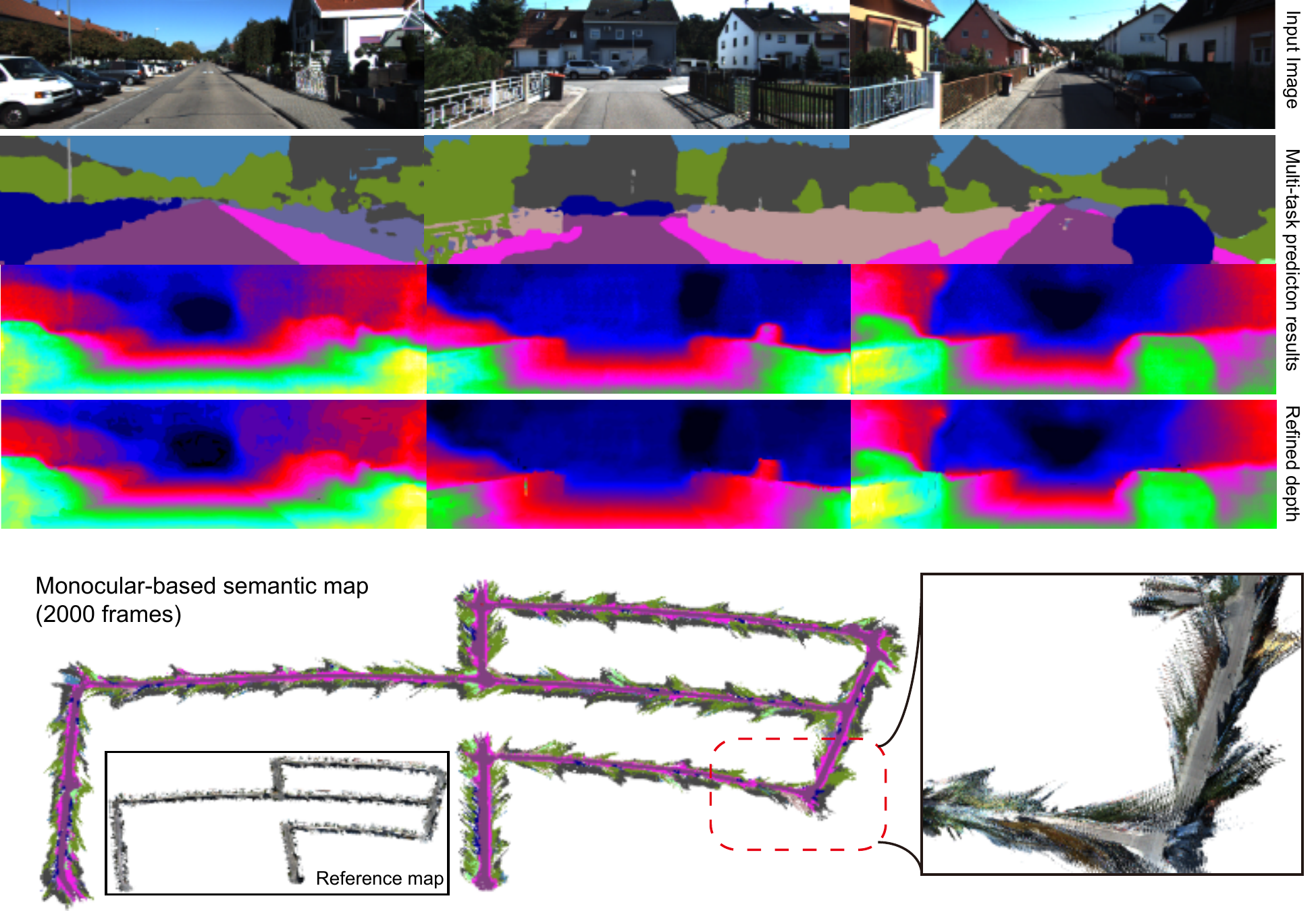}
	\caption{The 3D semantic reconstruction of proposed method. The first row demonstrates the input image while the following two rows are outputs of our network. The fourth row is the depth after refinement. The semantic map with 2000 monocular images from KITTI odometry dataset~\cite{geiger2012we} is displayed to prove the effectiveness.}
	\label{fig:demonstration}
\end{figure}

Recent visual Simultaneous Localization and Mapping (SLAM) systems could provide a sparse set of 3D points while positioning itself. The density of maps reconstructed from SLAM systems is not qualified for further demanding scene understanding. Stereo matching techniques require an additional camera to yield the disparity map. A recent stereo-based matching system~\cite{chen2017full} utilizes  cost estimation method for computing the similarity of image patches. Other methods, such as point cloud input~\cite{chen2016transforming}, will compute more slowly because there is too much data.   
Limited by the 3D representation method, the output map loses pixel-level accuracy. In this paper, with the depth and segmentation results from the proposed multi-task network, we further adopt the novel Simple Non-Iterative Clustering (SNIC) superpixelization method~\cite{achanta2017superpixels}~\cite{8500416} to reduce the inconsistency of the depth prediction result. 
We concerned the memory requirement of large-scale maps and save the final map by the vertices of superpixel units after polygonal partitioning. An example of our method is shown in Fig.~\ref{fig:demonstration}.

In the experiment, we quantitatively evaluate the results of both depth prediction and scene segmentation of proposed multi-task network on the Cityscape dataset~\cite{cordts2016cityscapes}. The reconstruction of large-scale maps is then conducted in the 3D space with monocular image streams from the KITTI benchmark~\cite{geiger2012we}. In both cases, our method performs better in accuracy as well as in terms of computational efficiency. The rest of the paper is organized as follows. We first review relative works on simultaneous prediction networks and monocular reconstruction in Section II, introduce the proposed method in Section III and demonstrate the experimental result in Section IV. The conclusion is stated in Section V.

\section{Related Work}
Depth prediction for scene understanding used to heavily rely on stereo vision~\cite{yamaguchi2014efficient,chen2017full}. Recent studies have been made progress in scene geometric understanding from the monocular camera. An encoder-decoder architecture~\cite{laina2016deeper} is proposed by Laina et al., which performs residual learning to predict dense depth maps . The new up-projection structure is adopted to avoid the checkerboard artifacts. Scene segmentation is another active field. The current leading segmentation network~\cite{chen2018encoder}, called Deeplab v3+, is based on their former work~\cite{chen2017rethinking} which could extract the boundaries unambiguously referring to the recovered structural information. 
For semantic reconstruction containing both depth and semantic prediction, separately applying these methods could bring higher computation cost and neglect the shared information between two tasks.

Multi-task learning techniques are designed to use the transfer feature between different tasks by jointly predict labels from a single model. Multi-task networks are adopted in the face attribute estimation, the contour detection, the semantic segmentation, etc. Neven et al. propose a network~\cite{neven2017fast} which combines scene segmentation, instance segmentation, and depth prediction into an integrated network based on the ENet. Kendall et al. use the Deeplab v3 as the fundamental architecture~\cite{kendall2017multi}. They conduct experiments with uncertainty weight losses demonstrating the superiority and effectiveness of homoscedastic uncertainty for multi-task learning.
However, these networks derive the same decoder architecture for each task ignoring the difference of outputs. In the proposed network, specialized decoders are designed to deal with different tasks, and connections between two decoders to transfer available information.

Semantic reconstruction can be basically divided into two categories. The first kind of methods are inheritors of 2D semantic segmentation results~\cite{kundu2014joint,barsan2018robust}. 
For monocular-based reconstructions, Kundu et al. propose an approach~\cite{kundu2014joint} to jointly infer geometric structure and 3D semantic labeling with a CRF model. 
The experimental results are good but with limitations. Due to the resolution and the ability of volumetric occupancy map, the output misses structural details. The second concerns the need to simultaneously provide semantic and geometric information in the 3D space. The incremental semantic reconstruction approach~\cite{vineet2015incremental} proposed by Vineet et al. builds the urban environment on a hash-based technique and a mean-field inference algorithm. The traditional stereo matching and visual odometry method are adopted to obtain basic 3D knowledge. 

\begin{figure*}[t]
	\centering
	\includegraphics[width=1.0\linewidth]{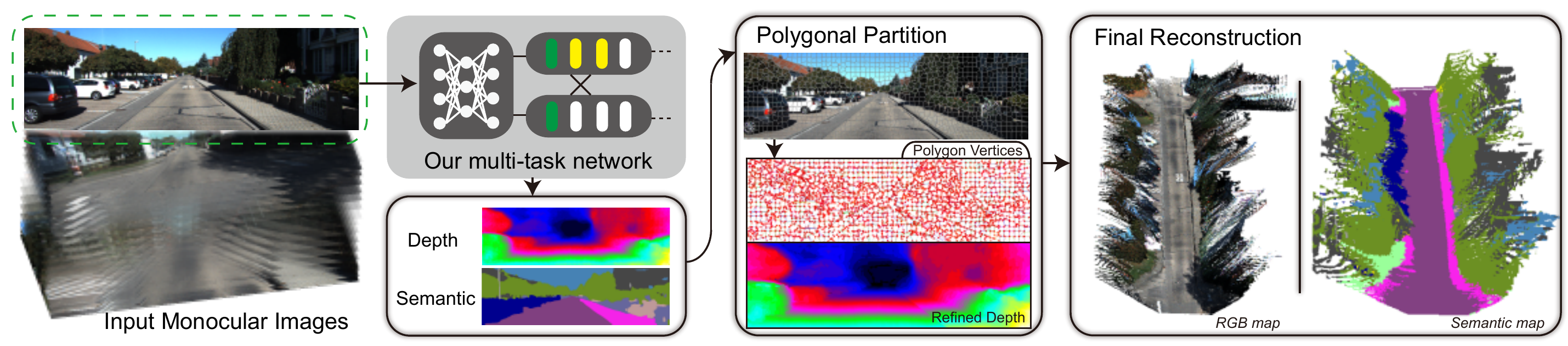}
	\caption{Overview of our monocular-based 3D reconstruction method. Given an image stream, the proposed multi-task network predicts its depth and semantic simultaneously. The polygonal superpixel segmentation is applied to refine the depth output while reducing the map memory cost. With the camera pose knowledge, we give the final 3D semantic maps rendering with RGB data and semantic information, respectively.}
	\label{fig:pipeline}
\end{figure*}

Superpixel segmentation~\cite{achanta2017superpixels} has been applied to promote stereo matching results. The matching algorithm of Yamaguchi et al.~\cite{yamaguchi2014efficient} called the SPS-st, whose formulation is based on the slanted-plane model with plain-fitting technique. 
We adopt this strategy to our depth prediction outputs with additional semantic knowledge which gives higher precision to boundaries during segmentation. To reduce the memory of large-scale reconstruction while maintaining pixel-level details, the map is stored with the vertices of superpixel after changedintopolygons~\cite{achanta2017superpixels}. The advancement of depth refinement and representation method are also conducted in the experiment.

\section{Monocular camera-based semantic reconstruction}

As shown in Fig.~\ref{fig:pipeline}, the input to the proposed method is a monocular image stream. The disparity map and the semantic segmentation are precalculated from the proposed multi-task network, whose architecture will be introduced in the following part. We use a superpixel segmentation method with additional depth and semantic knowledge to perform smoothing to the original depth. Further depth refinement is obtained on the planar surface, i.e. the road, under supervised by the semantic segmentation result. Each superpixel is then regarded as the basic unit for map storage after transforming into polygons, which saves memory requirement dramatically especially for large-scale structural space.

\subsection{Multi-task Network for Joint Inferring}
\begin{figure}[t]
	\centering
	\includegraphics[width=1.0\linewidth]{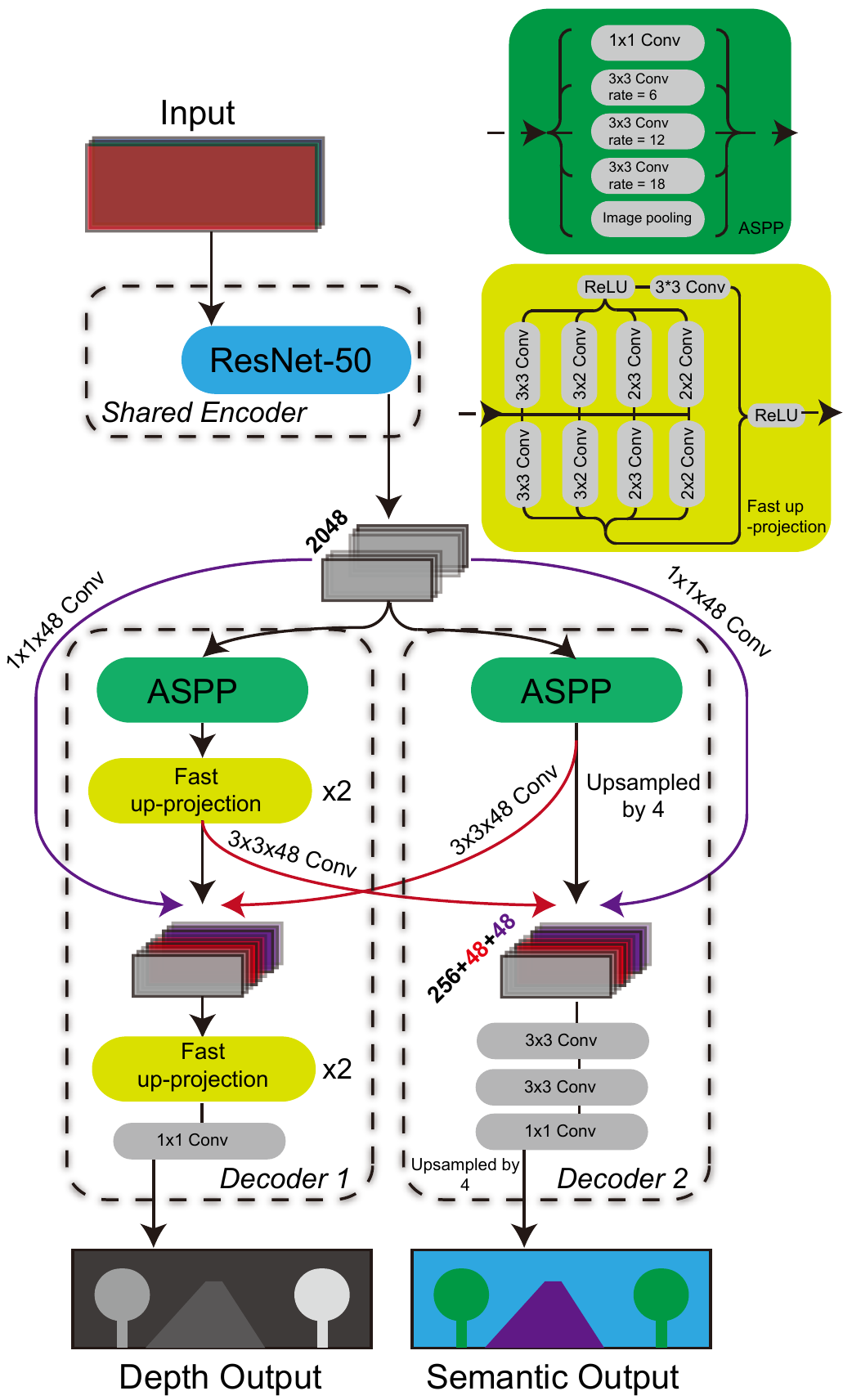}
	\caption{The proposed network architecture. Given an input RGB image, the network simultaneously produces the depth and the semantic prediction results. The detailed architecture of ASPP and fast up-projection display on the up-right corner.}
	\label{fig:network}
\end{figure}
The proposed network is demonstrated in Fig.~\ref{fig:network}. The network contains a shared encoder and two decoders. For the shared encoder, we use the ResNet-50~\cite{he2016deep} to produce rich and contextual features. The ResNet-50 is a trade-off choice between accuracy and memory requirements compared to the ResNet-101. Both decoders receive the same feature map from ResNet-50 as input while using ASPP modules to promote contextual awareness.The reason why we apply ASPP module in both decoders is that the decoders require different features. The semantic one deduces the categories by surrounding features and the global information is more useful for depth estimation. By applying filters of multiple rates to probe input feature maps, the network enlarges its fields-of-view to capture objects or image context of different scales.

The ASPP module in the depth decoder is to capture the discrepancy between nearby features to finally convert into depth. In the following upsampling process, we choose the fast up-projection structure~\cite{laina2016deeper} , which calculates group of filters with $3\times 3$, $2\times 2$, $2\times 3$ and $2\times 2$ convolutions on the feature map and interleaves them into a unified output matching the upsampling shape. Compared to deconvolutions, the interleaving mechanism helps to reduce these noises during depth prediction. The detailed difference is discussed in the experiment. We use the scale-invariant error~\cite{eigen2014depth} as the training loss for the depth prediction part, which adapts to multi-scale objects. The loss formulation $L_{d}$ is defined as
\begin{equation}
L_{d}(\tilde{y}_{d},y_{d})=\frac{1}{n}\sum_{i}^{n}d_{i}^{2}-\frac{1}{2n^{2}}(\sum_{i}^{n}d_{i})^{2},
\end{equation}
where $\tilde{y}_{d}$ and $y_{d}$ denote the depth prediction and the depth ground truth, respectively, $n$ is the volume of valid pixels and $d_{i}=\log \tilde{y}_{d,i}-\log y_{d,i}$.

In the semantic decoder, we connect the depth feature with a $3\times 3 \times 48$ convolution to provide useful information. The pixel sharing the same depth value or gradient has higher possibility to belong to the same semantic category. The cross-entropy loss is adopted to learn pixelwise label probabilities by averaging the loss over the pixels in each batch. The loss function $L_{s}$ for semantic prediction is
\begin{equation}
L_{s}(\tilde{y}_{s},y_{s})=\sum_{i}^{m}(-{y}_{s,i}\log{\tilde{y}_{s,i}}),
\end{equation}
where $\tilde{y}_{s}$ is the semantic prediction, $y_{s}$ is the corresponding ground truth and $m$ is the sum of classes. We minimize the combined loss, which is $L=\alpha L_{s}(\tilde{y}_{s},y_{s})+(1-\alpha)L_{d}(\tilde{y}_{d},y_{d})$, during our multi-task training process. The $\alpha$ is a balancing factor and is set to $0.75$ in our experiments.

\noindent\textbf{Training:} The training process is started form the ResNet-50 model pretrained on the ImageNet-1K dataset. Following the training protocol of Deeplab v3+~\cite{chen2018encoder}, we employ the same learning rate schedule, i.e., the ``poly" policy. The learning rate is set to $0.007$ while crop size is $512\times 1024$ for the Cityscape dataset~\cite{cordts2016cityscapes} and $192\times 624$ for the KITTI dataset~\cite{geiger2012we}. We use a random image scale and flip for data augmentation. In order to maintain the output of both decoders in the same interval, the possibility of semantic prediction belongs to $[0,1]$, and we transform the disparity label to log space divided by the maximum value. The proposed model is trained on the fine-annotated Cityscape dataset~\cite{cordts2016cityscapes} with $1575$ images containing both semantic and depth ground truths. $19$ labels are chosen as the training labels from $34$ official categories. To further conduct on the KITTI dataset~\cite{geiger2012we}, the network is trained with $200$ groups of images provided by the KITTI dataset.

\subsection{Polygonal Superpixel Partitioning}
Given the input image $\mathcal{I}$, we adopt the superpixel segmentation method to provide basic partition results $S=\{S_1, \dots, S_k\}$. The image $\mathcal{I}$ is transformed into the CIELAB color space. After initializing the centroid of pixels belonging to a regular grid, we use the following formulation to measure the distance between $k$th centroid $C[k]$ and $j$th candidate pixel, which is defined as
\begin{equation}
d_{j,k}=\sqrt{\frac{\|\textrm{x}_{j}-\textrm{x}_{k}\|_{2}^{2}}{s}+\frac{\|\textrm{c}_{j}-\textrm{c}_{k}\|_{2}^{2}}{m}+h(\tilde{y}_{s,j},\tilde{y}_{s,k})},
\end{equation}
where $s$ and $m$ are two normalizing factors for spatial and color distances, and $h(.,.)$ is set to $0$ when two pixels sharing the same semantic label and a larger value otherwise. We follow the simple non-iterative clustering~\cite{achanta2017superpixels} to update the boundary of superpixels, which requires less memory and performs faster .

With the superpixel segmentation result, we perform boundary tracing to each superpixel and transform superpixels into polygons. The vertices are then stored to represent each superpixel which saves memory compared to all boundary pixels. For the outdoor scene, especially for the driving environment, it is mainly composed of structural elements, such as the building and the road. Changing into polygons has an acceptable influence on the accuracy of boundaries.

\subsection{Depth Refinement}
The reason why we concerned about the smoothness of depth estimation is the uneven depth will lead to errors in perception system. The well-adopted assumption is made that pixels located in the same superpixel are belonging to the same depth plane surface. With the depth prediction $\tilde{y}_{d}$, the plane-fitting is performed to each superpixel to eliminate noises or fluctuations of the initial prediction. For the superpixel $S_{i}$, the plane parameter is defined as $(a_{i}, b_{i}, c_{i})$, which means the final depth for pixel $p(u,v)$ in $S_{i}$ is calculated as $a_{i}u+b_{i}v+c_{i}$. After refinement, we further smooth the road surface with random sample consensus algorithm. Note the depth refinement can be performed on other network prediction results. The improvement of depth is demonstrated in the experiment.

\subsection{Large-scale Reconstruction}
After producing the polygonal partition results with a plane function assigned to each polygon, we project polygons into the world coordinate. The projection parameter follows the corresponding training data, and we derive the camera pose from the visual odometry method~\cite{mur2015orb} or the available ground truth. Storing vertices requires much less memory compared to point clouds while maintaining comparable pixel-level accuracy. During reconstruction, we give each polygon a semantic label with the multi-task network prediction result.

\begin{table}[t]
	\scriptsize
	\caption{The results of our method on the Cityscape benchmark~\cite{cordts2016cityscapes}. We evaluate different decoder architectures. The term \textbf{deconv.}, \textbf{bi. interp.}, and \textbf{fast up-proj.} denote deconvolution layers, the bilinear interpolation and the fast up-projection module, respectively, and \textbf{RC} is the residual connection between decoders.}
	\begin{tabular}{cc|c|c}
		\hline
		\multicolumn{2}{c}{\multirow{2}{*}{Loss}}&\multicolumn{2}{|c}{Depth Estimation}\\
		&&{Mean Error$[px]$}&{RMS Error$[px]$}\\
		\hline
		{\multirow{5}{*}{Decoder}}&{\textit{\textbf{deconv.}}}&{$2.103$}&{$4.322$}\\
		\cline{2-4}
		&{\textit{\textbf{bi. interp.}}}&{$2.391$}&{$5.132$}\\
		\cline{2-4}
		&{\textit{\textbf{fast up-proj.}}}&{$1.913$}&{$4.001$}\\
		\cline{2-4}
		&{\textit{\textbf{fast up-proj.}}+\textit{\textbf{RC}}}&{$1.793$}&{$3.851$}\\
		\cline{2-4}
		&{\textit{\textbf{fast up-proj.}}+\textit{\textbf{RC}}+\textit{\textbf{refinement}}}&{$1.611$}&{$3.423$}\\
		\hline
	\end{tabular}
	\label{tab::decoder}
\end{table}

\begin{figure*}[!t]
	\centering
	\subfigure[]{
		\includegraphics[width=0.18\linewidth]{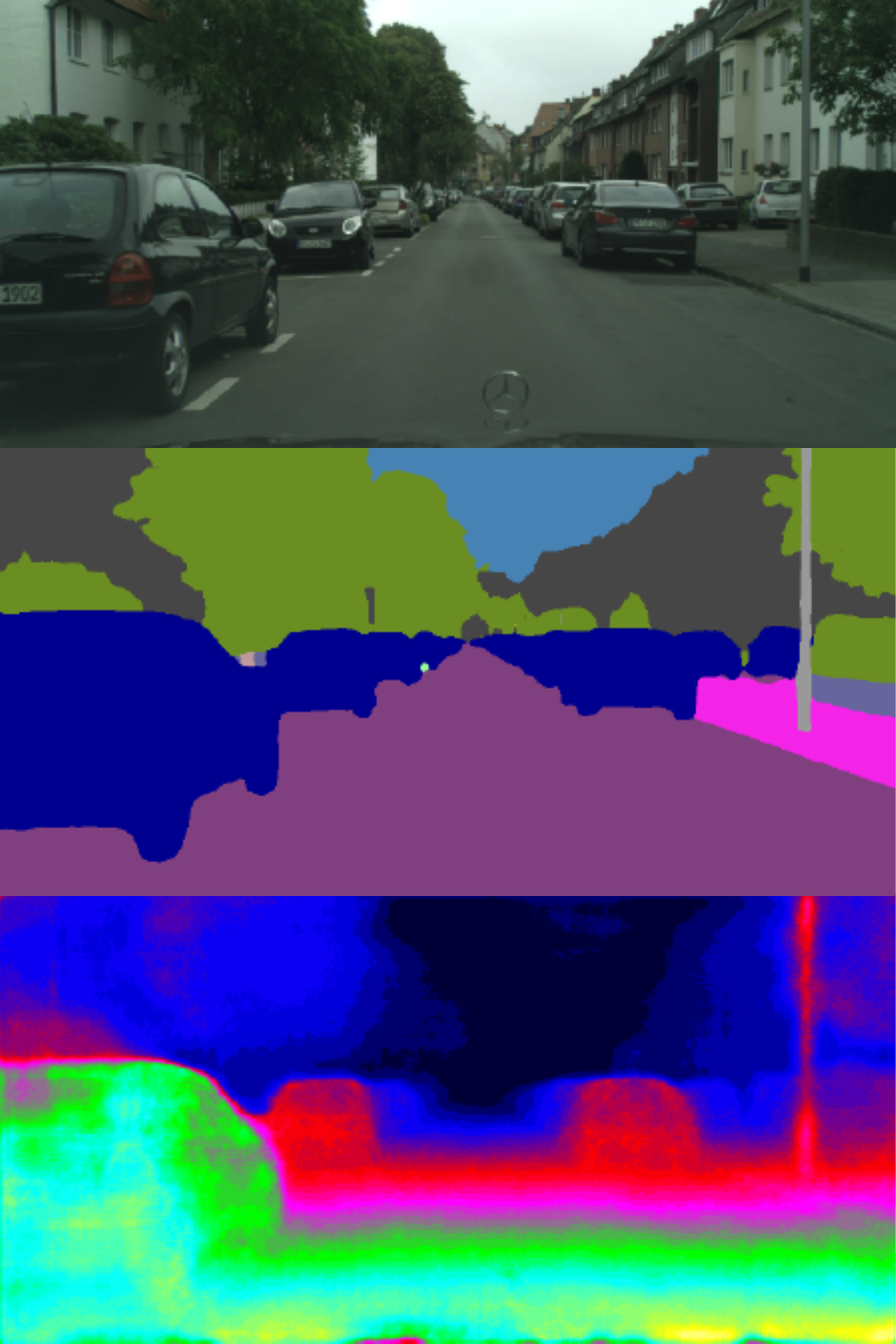}
	}
	\subfigure[]{
		\includegraphics[width=0.18\linewidth]{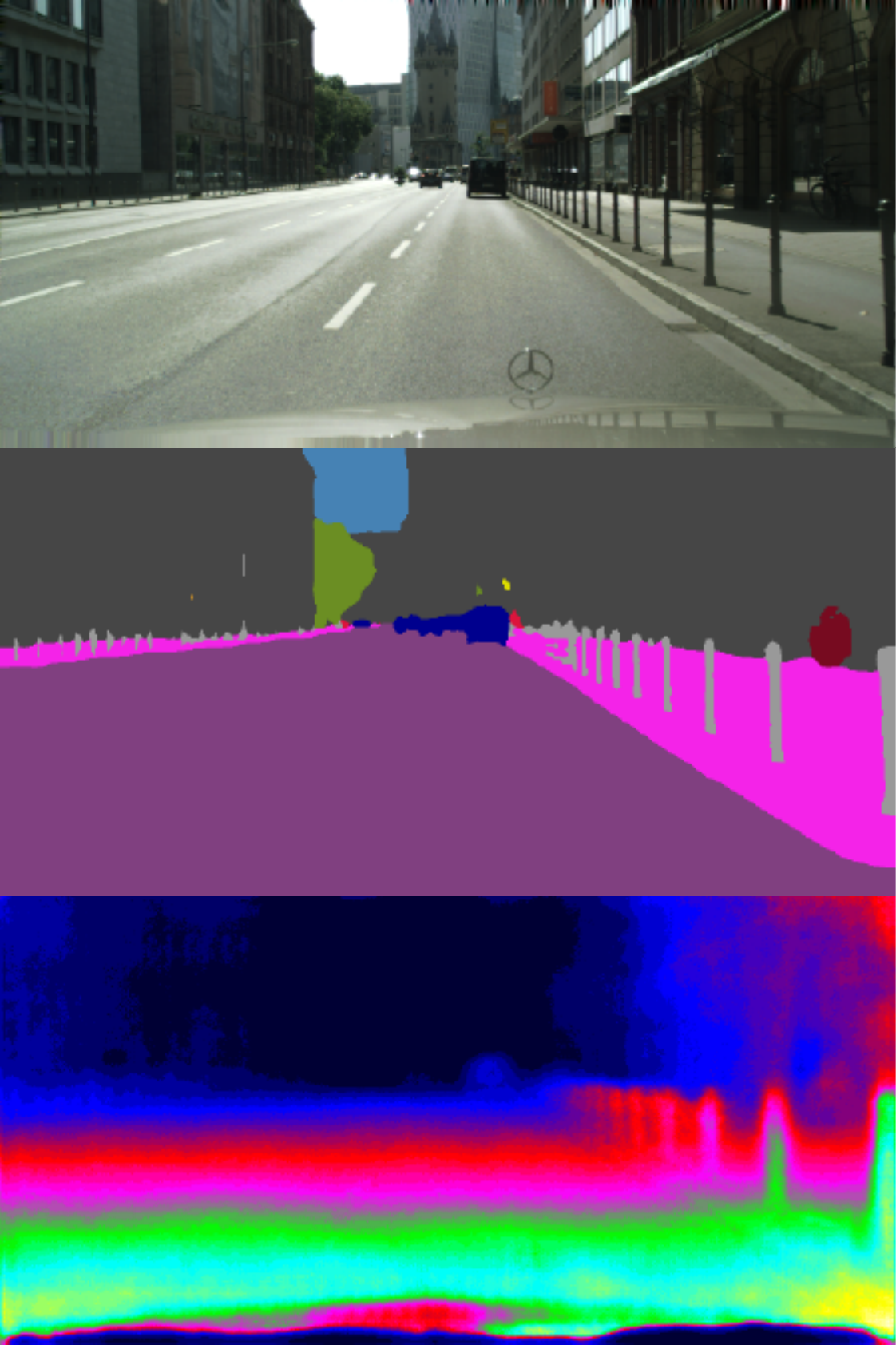}
	}
	\subfigure[]{
		\includegraphics[width=0.18\linewidth]{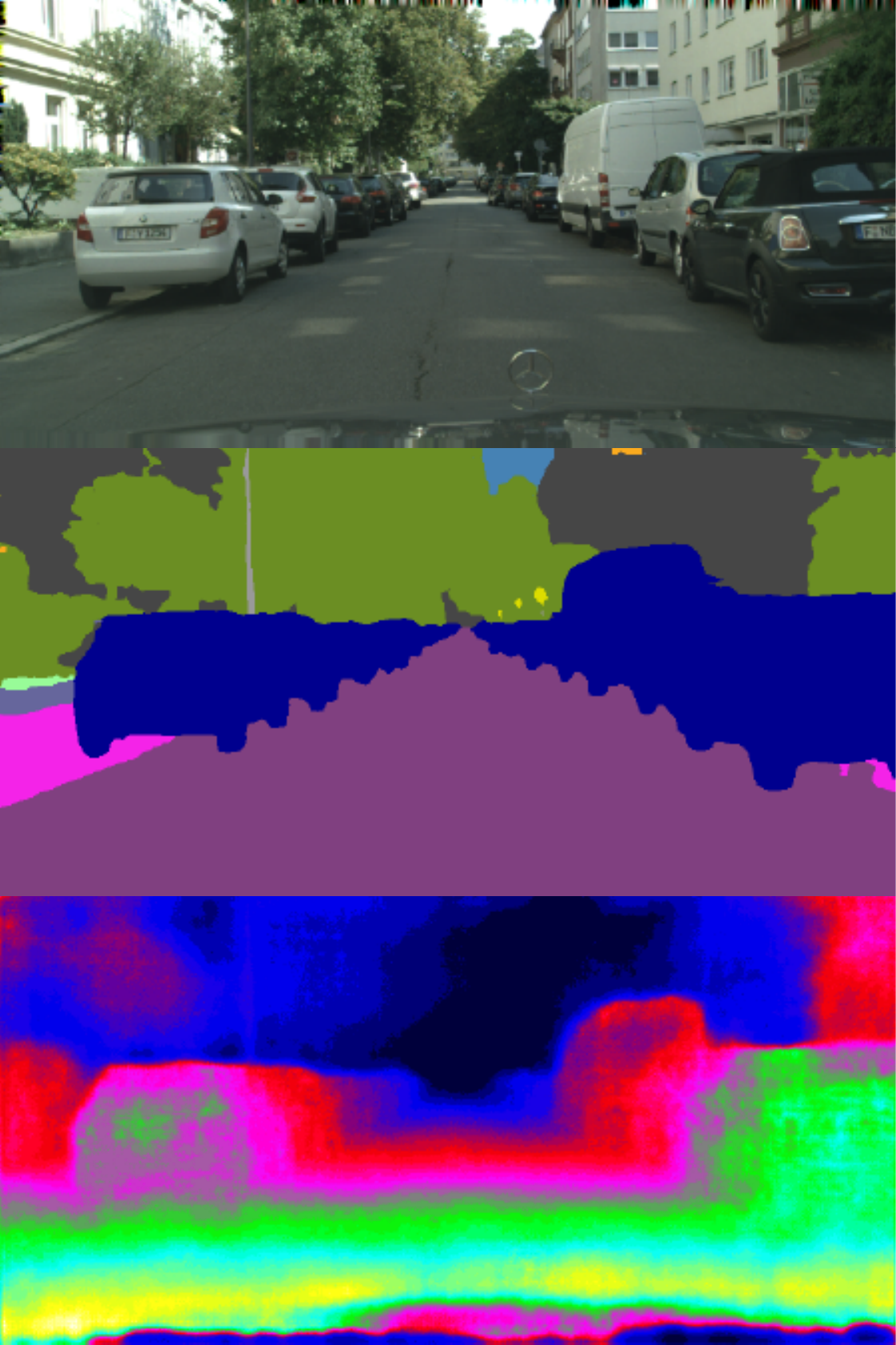}
	}
	\subfigure[]{
		\includegraphics[width=0.18\linewidth]{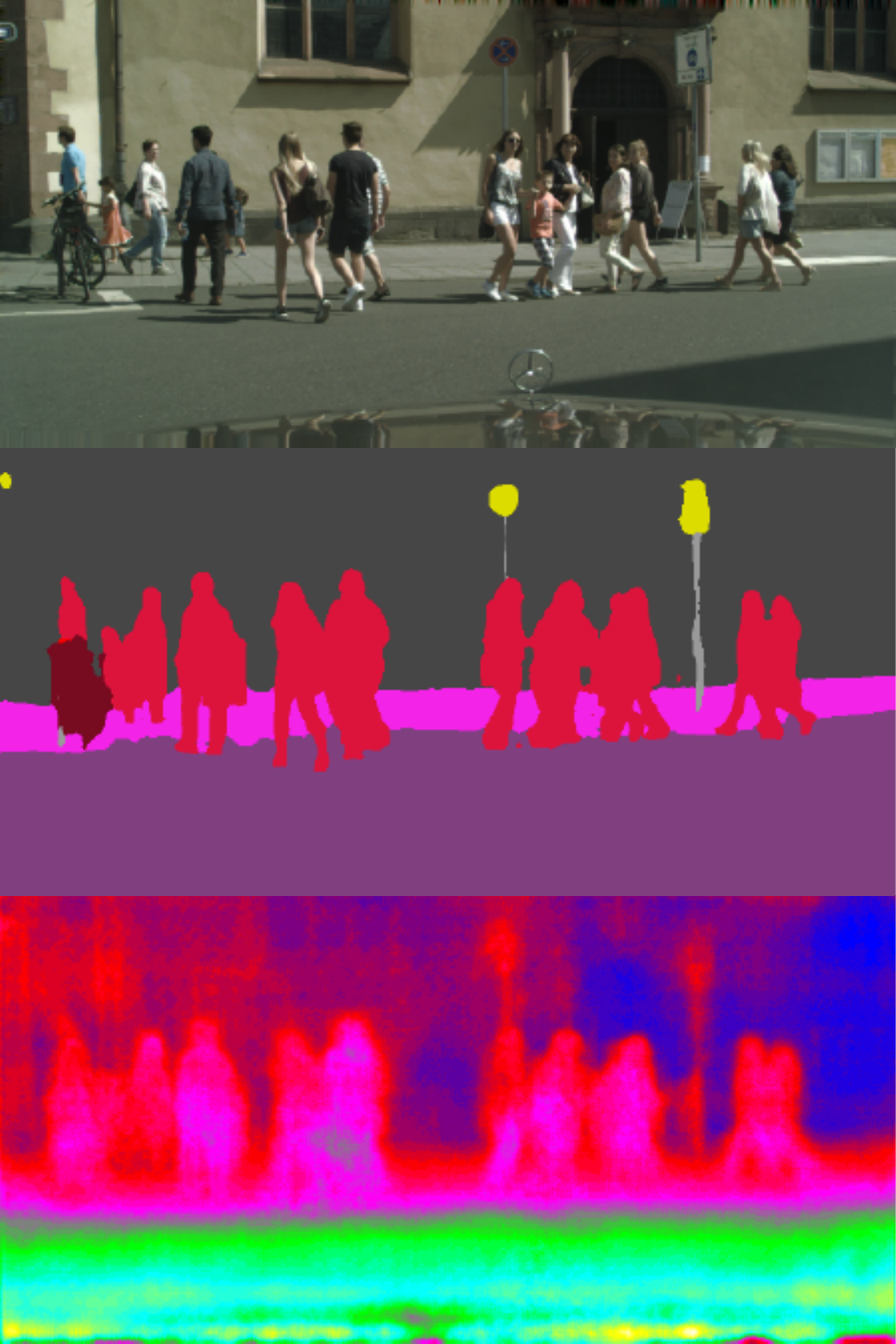}
	}
	\subfigure[]{
		\includegraphics[width=0.18\linewidth]{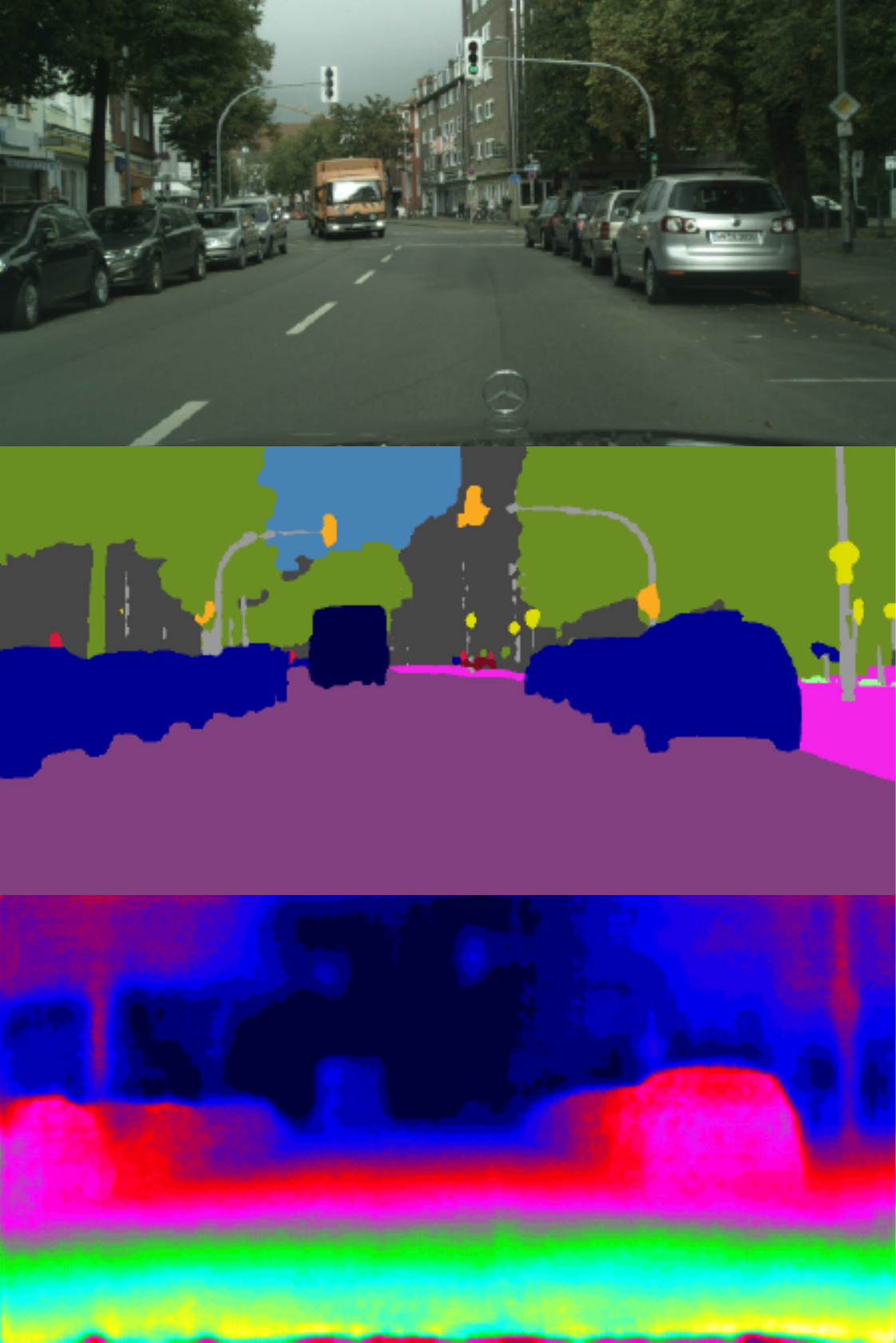}
	}
	
	\caption{The subfigure (a)-(e) are prediction result on the Cityscape validation dataset~\cite{cordts2016cityscapes}. Each group of result contains the input image, the semantic prediction and the depth prediction from top to bottom.}
	\label{fig::result1}
\end{figure*}

\begin{table*}[t]
	\label{tab::multitask}
	\centering
	\scriptsize
	\caption{The results on the Cityscape benchmark~\cite{cordts2016cityscapes}. We separate methods into three categories by their outputs. Note the comparison is not entirely fair, as many methods use ensembles of different training datasets and training image sizes.}
	\begin{tabular}{c|c|c|c|c|c|c|c|c|c}
		\hline
		\multirow{2}{*}{Loss}&\multicolumn{2}{|c|}{Segmentation}&\multicolumn{7}{|c}{Depth Estimation}\\
		&{IoU Cla.}&{IoU Cat.}&{Mean Error$[px]$}&{Abs Rel}&{RMS Error$[px]$}&{Sq Rel}&{$\delta<1.25$}&{$\delta<1.25^2$}&{$\delta<1.25^3$}\\
		\hline
		\multicolumn{10}{c}{\textbf{Monocular depth prediction only methods}}\\
		\hline
		{Zhou et al.~\cite{zhou2017unsupervised}}&{-}&{-}&{-}&{$0.267$}&{$7.85$}&{$2.686$}&{$0.577$}&{$0.840$}&{$0.937$}\\
		\hline
		{Kumar et al.~\cite{kumar2018monocular}}&{-}&{-}&{-}&{$0.393$}&{$10.50$}&{$4.683$}&{$0.352$}&{$0.689$}&{$0.905$}\\
		\hline
		\multicolumn{10}{c}{\textbf{Semantic segmentation only methods}}\\
		\hline
		{Deeplab v3+ (ResNet-101)~\cite{chen2018encoder}}&{$0.821$}&{$0.920$}&{-}&{-}&{-}&{-}&{-}&{-}&{-}\\
		\hline
		{PSPNet~\cite{zhao2017pyramid}}&{$0.802$}&{$0.906$}&{-}&{-}&{-}&{-}&{-}&{-}&{-}\\
		\hline
		\multicolumn{10}{c}{\textbf{Joint depth and semantic prediction methods}}\\
		\hline
		{Kendall et al.~\cite{kendall2017multi}}&{$0.785$}&{$0.899$}&{$2.920$}&{-}&{$5.880$}&{-}&{-}&{-}&{-}\\
		\hline
		{Neven et al.~\cite{neven2017fast}}&{$0.593$}&{$0.804$}&{-}&{-}&{-}&{-}&{-}&{-}&{-}\\
		\hline
		{Our Method (ResNet-50)}&{$0.708$}&{$0.879$}&{$1.793$}&{$0.220$}&{$3.851$}&{$2.310$}&{$0.655$}&{$0.851$}&{$0.933$}\\
		\hline
	\end{tabular}
\end{table*}

\section{Experiments}
The experiment is conducted in two parts. In the first part, we compare the proposed multi-task model performance with other networks~\cite{zhou2017unsupervised,kumar2018monocular,chen2018encoder,zhao2017pyramid,kendall2017multi,neven2017fast} in both semantic and depth prediction. The influence of adding connections between decoders and up-projection module is also demonstrated. The semantic decoder uses the common structures, so we don't evaluate it specifically. The depth results after refinement are also shown in this part. We evaluate our model on the Cityscape dataset~\cite{cordts2016cityscapes}, which aims at road scene understanding tasks. Qualitative results on the large-scale dataset, namely the KITTI odometry dataset~\cite{geiger2012we}, are displayed in the second part to show the effectiveness of proposed monocular reconstruction method. The urban scene provided by the KITTI benchmark is quite challenging, as it covers a large area and contains significant depth variation. 

\subsection{Multi-task Network}

\begin{figure}[t]
	\centering
	\subfigure[]{
		\includegraphics[width=0.29\linewidth]{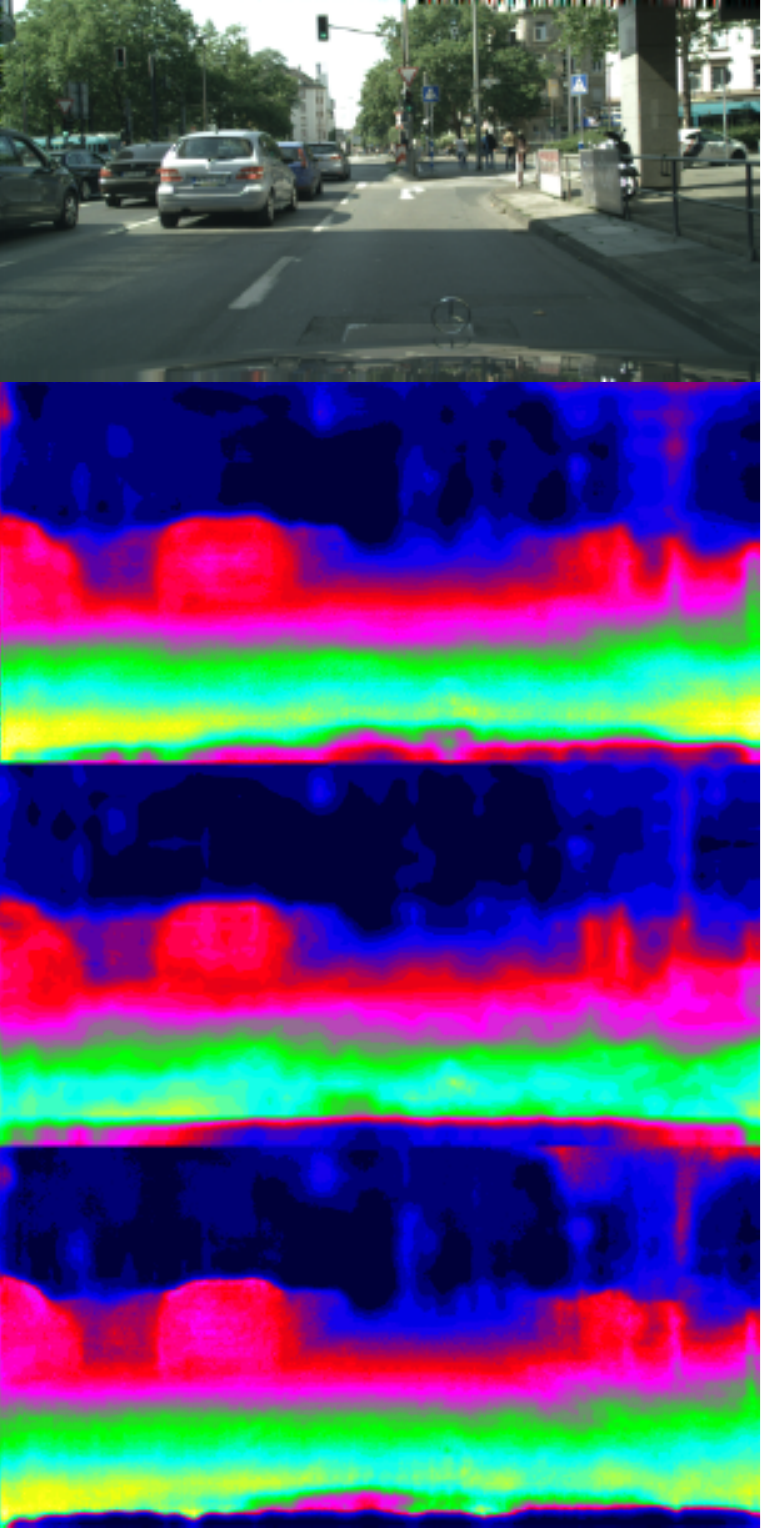}
	}
	\subfigure[]{
		\includegraphics[width=0.29\linewidth]{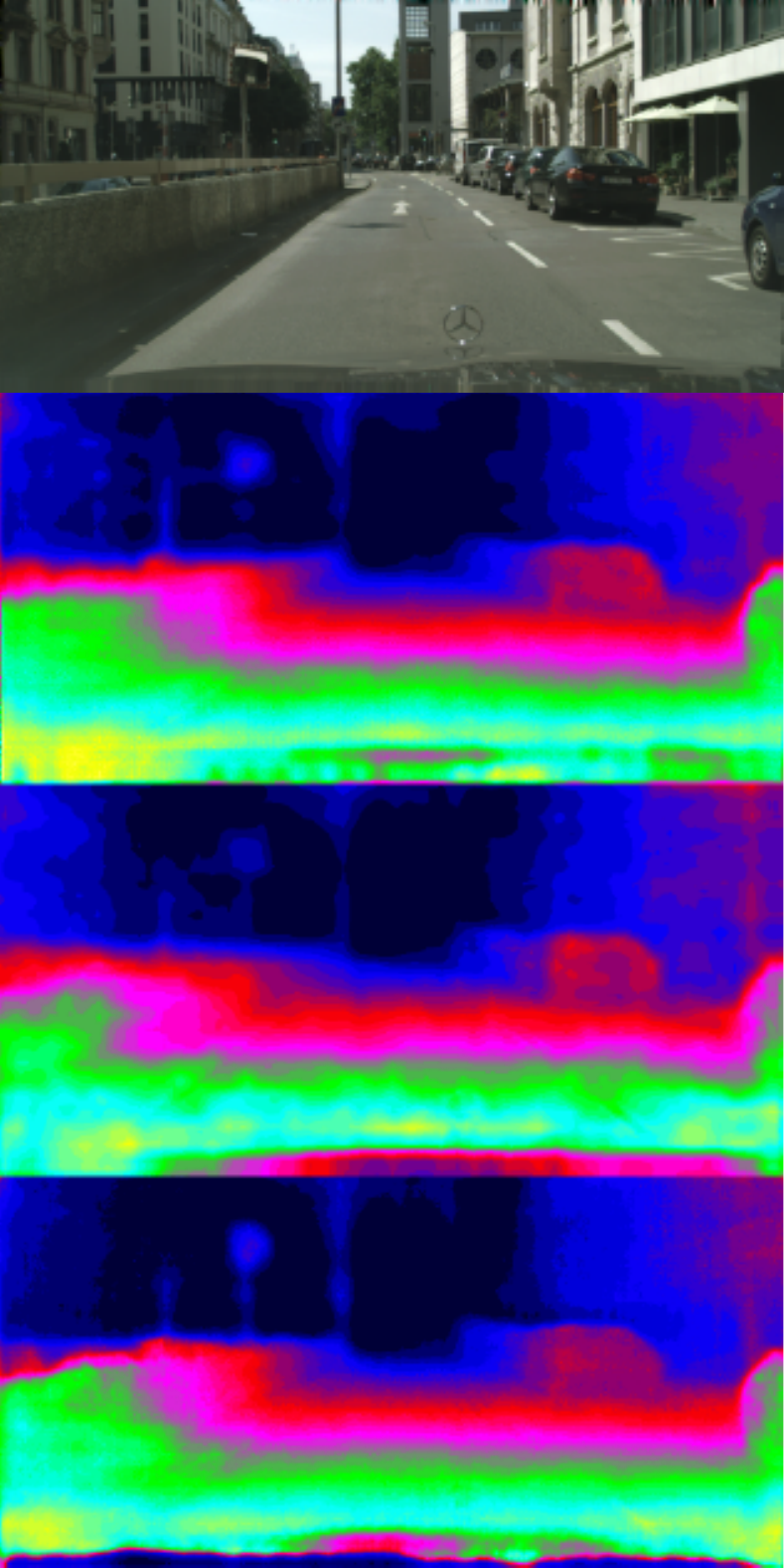}
	}
	\subfigure[]{
		\includegraphics[width=0.29\linewidth]{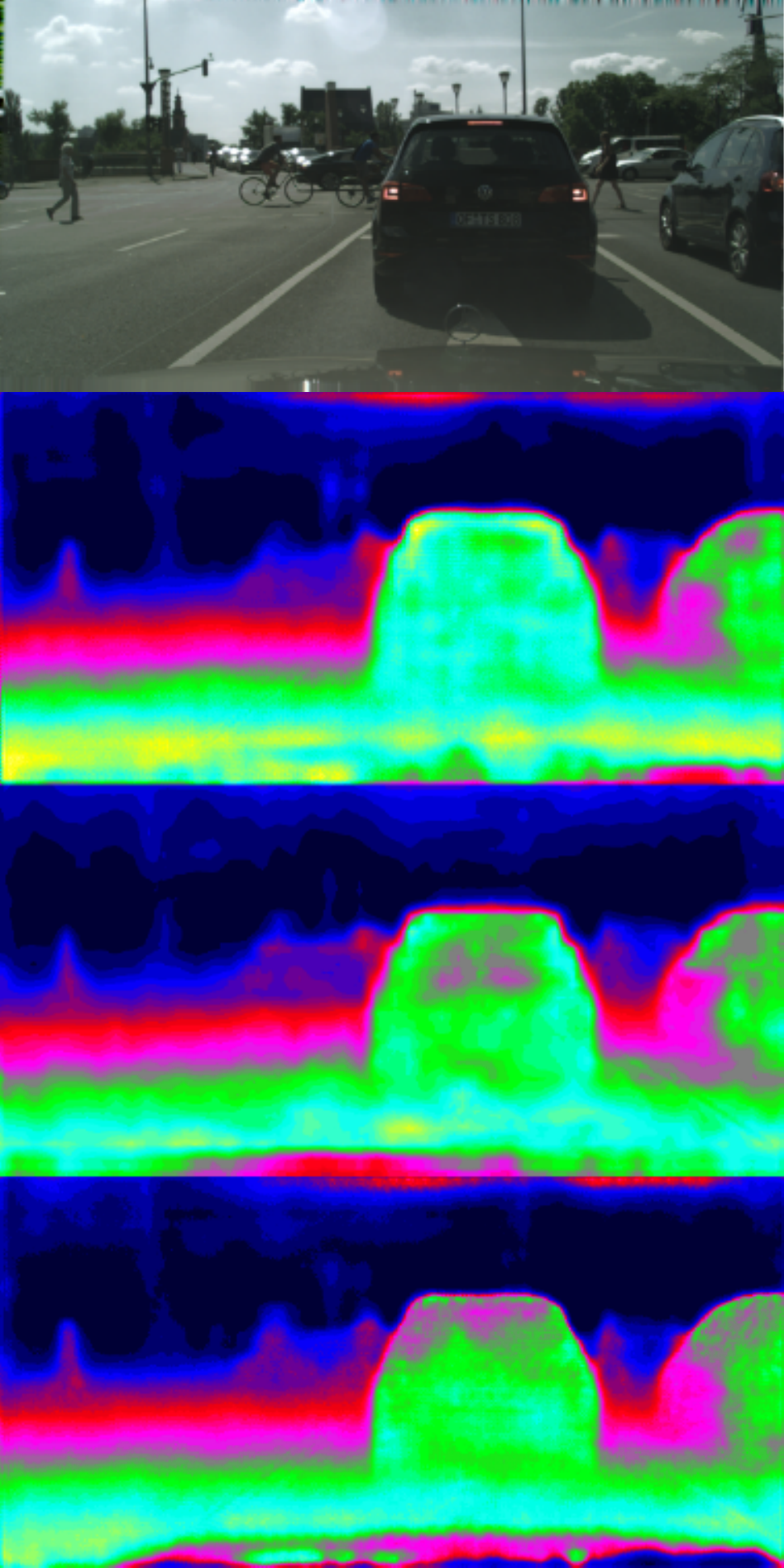}
	}
	\caption{Comparison after modifying the decoder architecture. For each group of depth prediction, we separately use the deconvolution layer, the bilinear interpolation, and the fast up-projection from top to bottom for recovering to the original size.}
	\label{fig::result_compare}
\end{figure}

The performance of proposed network is evaluated in two different aspects. We first change the upsampling component in our depth decoder into the deconvolution, the bilinear interpolation, and the fast up-projection module utilized in this paper. The three modules upsample the feature map into the output with target sizes. As shown in Fig.~\ref{fig::result_compare}, the blur effect heavily occurs in the output with the decoder of deconvolution. Checkerboard effects is easy to notice especially for planar surfaces, such as the road. Due to the mechanism of bilinear interpolation which is widely used in semantic prediction networks, the uneven disparity happens. The accuracy of different decoder architectures are displayed in Tab.~I. Using the residual connection and further depth refinement with superpixels also improves the performance. Examples of our results on the Cityscape validation set~\cite{cordts2016cityscapes} is shown in Fig.~\ref{fig::result1}.

We then compare to a number of related works~\cite{zhou2017unsupervised,kumar2018monocular,chen2018encoder,zhao2017pyramid,kendall2017multi,neven2017fast} in Tab.~II. Three categories of methods are demonstrated. In terms of semantic segmentation, our model lags behind the Deeplab v3+. The reason is partly due to the encoder difference, i.e., the ResNet-50 contains much fewer parameters and is shallower than the ResNet-101. For depth prediction, the proposed method clearly outperforms other models. The detailed definition of loss metric Abs Rel, Sq Rel, and etc. follows ~\cite{eigen2014depth}. The higher of $\delta<1.25, 1.25^2, 1.25^3$ denote better performance, and the other metrics are on the contrary.

\subsection{Reconstruction on the KITTI Dataset}

\begin{figure}[t]
	\centering
	\subfigure[]{
		\includegraphics[width=1.0\linewidth]{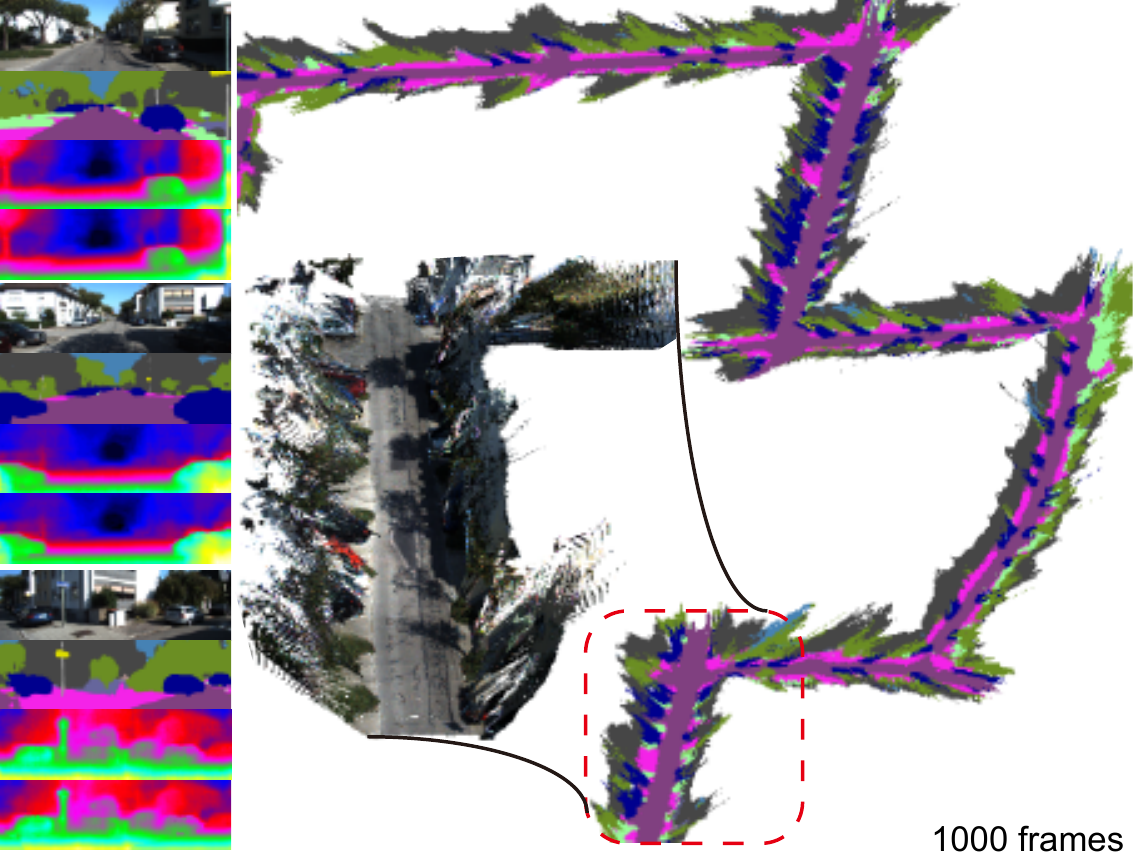}
	}
	\subfigure[]{
		\includegraphics[width=1.0\linewidth]{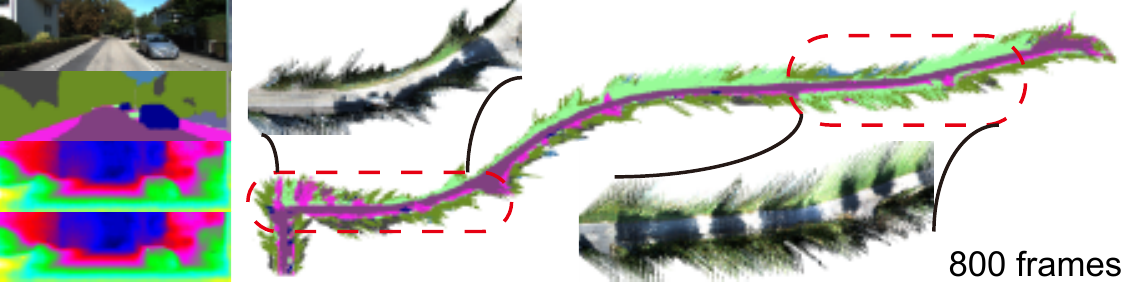}
	}
	\caption{Two demonstrations of our 3D semantic map on long outdoor image sequences. We also zoom in and textured with RGB data to illustrate details. The example group of semantic and depth prediction are displayed on the left Eide. The depth after refinement is at the last row of each group.}
	\label{fig::result_map}
\end{figure}

 The qualitative experiments of outdoor scenes are based on the KITTI odometry dataset~\cite{geiger2012we}. In this part, we intend to show the 3D semantic reconstructions on long monocular image sequences covering a large urban area. Due to few semantic annotations of KITTI dataset, we generated the semantic results by pretrain model on Cityscapes. The camera pose follows the provided ground truth. As demonstrated in Fig.~\ref{fig::result_map}, we choose the sequence $0$ and $3$ from KITTI odometry dataset with $1000$ frames, $800$  frames, respectively. In the first example (Fig.~\ref{fig::result_map} (a)), the vehicle located on the roadside is presented due to correct semantic segmentation. With the depth information, the position of vehicles in the 3D world can be obtained. The key aspect of the proposed 3D map is to convey the height information which could assist road area extraction, which is well-displayed in the RGB map. The large-scale map with semantic information is displayed with polygons in the 3D space, which saves the required memory space. Another map of a countryside road with two RGB textured aerial views is shown in Fig.~\ref{fig::result_map} (b). We provide groups of semantic and depth prediction on the left side with the depth after refinement displayed at the last row. The superpixel  helps extract the boundary of objects, which eliminates the  blur effect in the depth of the network output.

\section{CONCLUSIONS}
In this paper, we presented a method to reconstruct dense large-scale semantic maps based on a monocular camera. The semantic and depth are first predicted jointly by a novel multi-task network. The proposed network improves results with ASPP modules enlarging the fields-of-view and the fast up-projection in decoder reducing checkerboard artifacts. The depth prediction also yields data from the semantic decoder with the residual connection. The superpixels are extracted with a depth- and semantic-aware manner and further changed into polygons. Inconsistencies, i.e., the fluctuation in the depth prediction, are reduced after performing plane-fitting to each segment. The experimental results show the effectiveness.

\section*{ACKNOWLEDGE}
This work was supported in part by the National Natural Science Foundation of China under Grant 61773414 and under Grant 2018YFB1305002

\bibliographystyle{ieeetr}
\bibliography{ref}

\begin{thebibliography}{10}

\bibitem{chen2017moving}
L.~Chen, L.~Fan, G.~Xie, K.~Huang, and A.~N{\"u}chter, ``Moving-object
  detection from consecutive stereo pairs using slanted plane smoothing,'' {\em
  IEEE Transactions on Intelligent Transportation Systems}, vol.~18, no.~11,
  pp.~3093--3102, 2017.

\bibitem{li2014sensor}
Q.~Li, L.~Chen, M.~Li, S.-L. Shaw, and A.~N{\"u}chter, ``A sensor-fusion
  drivable-region and lane-detection system for autonomous vehicle navigation
  in challenging road scenarios,'' {\em IEEE Transactions on Vehicular
  Technology}, vol.~63, no.~2, pp.~540--555, 2014.

\bibitem{zhao2017pyramid}
H.~Zhao, J.~Shi, X.~Qi, X.~Wang, and J.~Jia, ``Pyramid scene parsing network,''
  in {\em IEEE Conf. on Computer Vision and Pattern Recognition (CVPR)},
  pp.~2881--2890, 2017.

\bibitem{chen2018encoder}
L.-C. Chen, Y.~Zhu, G.~Papandreou, F.~Schroff, and H.~Adam, ``Encoder-decoder
  with atrous separable convolution for semantic image segmentation,'' {\em
  arXiv preprint arXiv:1802.02611}, 2018.

\bibitem{eigen2014depth}
D.~Eigen, C.~Puhrsch, and R.~Fergus, ``Depth map prediction from a single image
  using a multi-scale deep network,'' in {\em Advances in neural information
  processing systems}, pp.~2366--2374, 2014.

\bibitem{laina2016deeper}
I.~Laina, C.~Rupprecht, V.~Belagiannis, F.~Tombari, and N.~Navab, ``Deeper
  depth prediction with fully convolutional residual networks,'' in {\em 3D
  Vision (3DV), 2016 Fourth International Conference on}, pp.~239--248, IEEE,
  2016.

\bibitem{kumar2018monocular}
A.~C. Kumar, S.~M. Bhandarkar, and M.~Prasad, ``Monocular depth prediction
  using generative adversarial networks,'' in {\em 1st International Workshop
  on Deep Learning for Visual SLAM,(CVPR)}, vol.~3, p.~7, 2018.

\bibitem{zhou2017unsupervised}
T.~Zhou, M.~Brown, N.~Snavely, and D.~G. Lowe, ``Unsupervised learning of depth
  and ego-motion from video,'' in {\em CVPR}, vol.~2, p.~7, 2017.

\bibitem{kendall2017multi}
A.~Kendall, Y.~Gal, and R.~Cipolla, ``Multi-task learning using uncertainty to
  weigh losses for scene geometry and semantics,'' {\em arXiv preprint
  arXiv:1705.07115}, vol.~3, 2017.

\bibitem{neven2017fast}
D.~Neven, B.~De~Brabandere, S.~Georgoulis, M.~Proesmans, and L.~Van~Gool,
  ``Fast scene understanding for autonomous driving,'' {\em arXiv preprint
  arXiv:1708.02550}, 2017.

\bibitem{caruana1997multitask}
R.~Caruana, ``Multitask learning,'' {\em Machine learning}, vol.~28, no.~1,
  pp.~41--75, 1997.

\bibitem{geiger2012we}
A.~Geiger, P.~Lenz, and R.~Urtasun, ``Are we ready for autonomous driving? the
  kitti vision benchmark suite,'' in {\em Computer Vision and Pattern
  Recognition (CVPR), 2012 IEEE Conference on}, pp.~3354--3361, IEEE, 2012.

\bibitem{chen2017full}
L.~Chen, L.~Fan, J.~Chen, D.~Cao, and F.~Wang, ``A full density stereo matching
  system based on the combination of cnns and slanted-planes,'' {\em IEEE
  Transactions on Systems, Man, and Cybernetics: Systems}, 2017.

\bibitem{chen2016transforming}
L.~Chen, Y.~He, J.~Chen, Q.~Li, and Q.~Zou, ``Transforming a 3-d lidar point
  cloud into a 2-d dense depth map through a parameter self-adaptive
  framework,'' {\em IEEE Transactions on Intelligent Transportation Systems},
  vol.~18, no.~1, pp.~165--176, 2016.

\bibitem{achanta2017superpixels}
R.~Achanta and S.~S{\"u}sstrunk, ``Superpixels and polygons using simple
  non-iterative clustering,'' in {\em Computer Vision and Pattern Recognition
  (CVPR), 2017 IEEE Conference on}, pp.~4895--4904, Ieee, 2017.

\bibitem{8500416}
L.~{Fan}, L.~{Chen}, K.~{Huang}, and D.~{Cao}, ``Planecell: Representing
  structural space with plane elements,'' in {\em 2018 IEEE Intelligent
  Vehicles Symposium (IV)}, pp.~978--985, June 2018.

\bibitem{cordts2016cityscapes}
M.~Cordts, M.~Omran, S.~Ramos, T.~Rehfeld, M.~Enzweiler, R.~Benenson,
  U.~Franke, S.~Roth, and B.~Schiele, ``The cityscapes dataset for semantic
  urban scene understanding,'' in {\em Proceedings of the IEEE conference on
  computer vision and pattern recognition}, pp.~3213--3223, 2016.

\bibitem{yamaguchi2014efficient}
K.~Yamaguchi, D.~McAllester, and R.~Urtasun, ``Efficient joint segmentation,
  occlusion labeling, stereo and flow estimation,'' in {\em European Conference
  on Computer Vision}, pp.~756--771, Springer, 2014.

\bibitem{chen2017rethinking}
L.-C. Chen, G.~Papandreou, F.~Schroff, and H.~Adam, ``Rethinking atrous
  convolution for semantic image segmentation,'' {\em arXiv preprint
  arXiv:1706.05587}, 2017.

\bibitem{kundu2014joint}
A.~Kundu, Y.~Li, F.~Dellaert, F.~Li, and J.~M. Rehg, ``Joint semantic
  segmentation and 3d reconstruction from monocular video,'' in {\em European
  Conference on Computer Vision}, pp.~703--718, Springer, 2014.

\bibitem{barsan2018robust}
I.~A. B{\^a}rsan, P.~Liu, M.~Pollefeys, and A.~Geiger, ``Robust dense mapping
  for large-scale dynamic environments,'' in {\em Proceedings of the IEEE
  International Conference on Robotics and Automation (ICRA)}, 2018.

\bibitem{vineet2015incremental}
V.~Vineet, O.~Miksik, M.~Lidegaard, M.~Nie{\ss}ner, S.~Golodetz, V.~A.
  Prisacariu, O.~K{\"a}hler, D.~W. Murray, S.~Izadi, P.~P{\'e}rez, {\em
  et~al.}, ``Incremental dense semantic stereo fusion for large-scale semantic
  scene reconstruction,'' in {\em Robotics and Automation (ICRA), 2015 IEEE
  International Conference on}, pp.~75--82, IEEE, 2015.

\bibitem{he2016deep}
K.~He, X.~Zhang, S.~Ren, and J.~Sun, ``Deep residual learning for image
  recognition,'' in {\em Proceedings of the IEEE conference on computer vision
  and pattern recognition}, pp.~770--778, 2016.

\bibitem{mur2015orb}
R.~Mur-Artal, J.~M.~M. Montiel, and J.~D. Tardos, ``Orb-slam: a versatile and
  accurate monocular slam system,'' {\em IEEE Transactions on Robotics},
  vol.~31, no.~5, pp.~1147--1163, 2015.

\end{thebibliography}

\end{document}